\documentclass{article}

\usepackage[final]{corl_2018}

\usepackage[utf8]{inputenc} 
\usepackage[T1]{fontenc}    
\usepackage{hyperref}       
\usepackage{url}            
\usepackage{booktabs}       
\usepackage{amsfonts}       
\usepackage{nicefrac}       
\usepackage{microtype}      

\usepackage{graphicx} 
\usepackage[font=footnotesize]{caption}
\usepackage{subcaption}
\usepackage{wrapfig}
\usepackage{rotating}
\usepackage{sidecap}
\usepackage{algorithm}
\usepackage{algorithmic}
\usepackage{amsmath}
\usepackage{amssymb}
\usepackage{xspace}
\usepackage{color}

\usepackage{multirow}

\newcommand{\bs}{\mathbf{s}}
\newcommand{\ba}{\mathbf{a}}
\newcommand{\bA}{\mathbf{A}}

\newcommand{\specialcell}[2][c]{%
  \renewcommand{\arraystretch}{0.5}\begin{tabular}[#1]{@{}c@{}}#2\end{tabular}\renewcommand{\arraystretch}{2.0}}

\renewcommand{\arraystretch}{1.5}
\newcommand\blfootnote[1]{%
  \begingroup
  \renewcommand\thefootnote{}\footnote{#1}%
  \addtocounter{footnote}{-1}%
  \endgroup
}

\newcommand\rurl[1]{%
  \href{http://#1}{\nolinkurl{#1}}%
}

\title{Composable Action-Conditioned Predictors:\\Flexible Off-Policy Learning for Robot Navigation}

\author{
  Gregory Kahn$^{*}$, Adam Villaflor$^{*}$, Pieter Abbeel, Sergey Levine\\
  Berkeley AI Research (BAIR), University of California, Berkeley
}

\begin{document}
\maketitle

\blfootnote{$^{*}$ The first two authors contributed equally to this work.}


\vspace*{-40pt}
\begin{abstract}
A general-purpose intelligent robot must be able to learn autonomously and be able to accomplish multiple tasks in order to be deployed in the real world. However, standard reinforcement learning approaches learn separate task-specific policies and assume the reward function for each task is known a priori. We propose a framework that learns event cues from off-policy data, and can flexibly combine these event cues at test time to accomplish different tasks. These event cue labels are not assumed to be known a priori, but are instead labeled using learned models, such as computer vision detectors, and then ``backed up'' in time using an action-conditioned predictive model. We show that a simulated robotic car and a real-world RC car can gather data and train fully autonomously without any human-provided labels beyond those needed to train the detectors, and then at test-time be able to accomplish a variety of different tasks. Videos of the experiments and code can be found at \rurl{github.com/gkahn13/CAPs}
\end{abstract}

\keywords{multi-task learning, rewards, prediction, robot navigation}


\section{Introduction}

\begin{wrapfigure}{r}{0.4\textwidth}
	\vspace*{-10pt}
	\centering
	\includegraphics[width=0.4\textwidth]{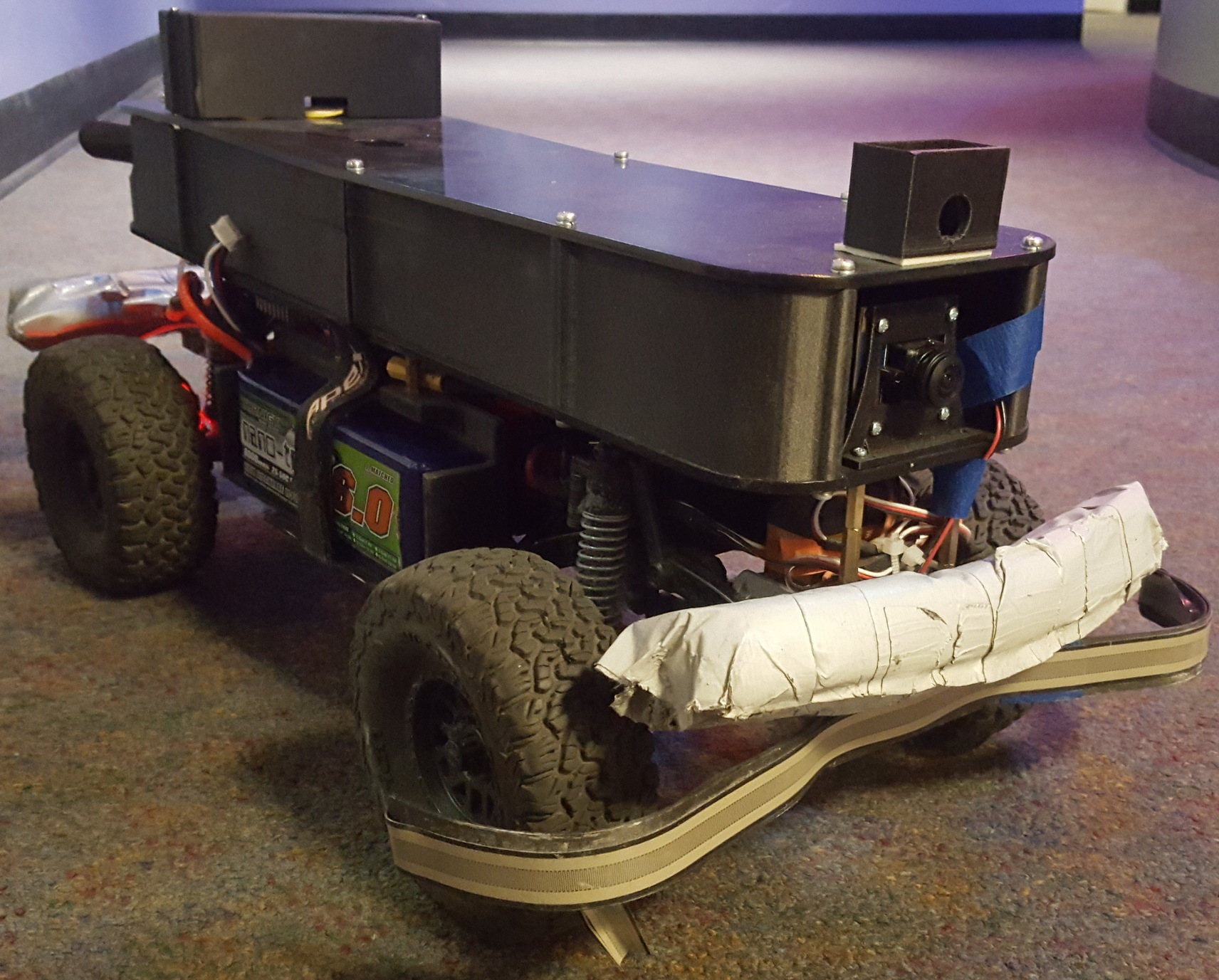}
	\caption{Our method can learn flexible mobility skills off-policy for a real-world RC car.}
	\label{fig:rccar-teaser}
	\vspace*{-12pt}
\end{wrapfigure}

A general-purpose intelligent agent, such as a mobile robot, must be able to perform a wide variety of tasks. Reinforcement learning provides a tool for learning such goal-directed behavior. However, standard reinforcement learning approaches aim to solve individual tasks with task-specific policies and assume that the task is defined by an extrinsic reward function that is provided \textit{a priori}. These assumptions can be too restrictive for real-world problems. First, the specifics of the task might vary: at one point in time, we might want a mobile robot to reach a user-specified destination as quickly as possible, while at another point, we might want the robot to search for an object in its environment. Learning separate policies for each task can quickly become impractical. Second, the cues that the robot can automatically extract from its environment for learning useful skills are limited to those that can be perceived autonomously: perfect extrinsic reward functions are an ideal assumption that is rarely satisfied in real-world settings.

We propose a generalization of the reinforcement learning framework that combines flexible multi-task learning, off-policy training, and the ability to learn directly from real-world events that can be detected automatically, for example with modern computer vision systems. The main idea behind our approach is to ``back up'' a set of event cues such that future values of those cues can be predicted based on the current observation (e.g., camera image) and actions. At test time, these predictors can be recombined to achieve user-specified goals. The user can flexibly specify the desired combination of events, such as what speed to move and which lane to drive in, and the robot evaluates candidate actions to determine which actions will maximize the likelihood of those events occurring in the future. We call our models composable action-conditioned predictors (CAPs).
From the standpoint of reinforcement learning, our method generalizes value function methods to predict multiple events, and then uses these event predictors to flexibly solve various tasks at test time. From the standpoint of robotic perception, our method can be viewed as a way to turn a vision system into a prediction system: the output of a vision system that \emph{detects} an event can be used to train a model that \emph{predicts} that event, allowing the robot to plan how to cause (or avoid) this event.

The main contribution of our work is CAPs, a general framework for multi-objective learning-based control that scales to large datasets, deep neural network function approximators, and off-policy training. We demonstrate that this framework can be used to train a robot to accomplish a variety of user-specified goals at test time. The individual event cues we consider include those that can be trivially labeled by the robot itself---such as collisions, speed, and heading---as well as event cues labelled by a learned detection system, such as road lanes and doorways. The framework is general and can accommodate any event cue. Our experimental evaluation consists of training these predictors for both a simulated car and real-world RC car (Fig.~\ref{fig:rccar-teaser}) using data collected entirely autonomously, without any human-provided labels beyond those needed to train the object detector. At test-time, we illustrate behaviors such as collision avoidance; path, road, and heading following; and speed control.



\section{Related Work}
While most reinforcement learning methods learn one task at a time~\cite{SuttonBarto}, a number of prior works have studied how multiple skills can be acquired simultaneously. Multi-task methods often aim to capture multiple tasks within the same policy or value function~\cite{barreto2017successor,wilson2007multi}. In contrast, our approach aims to learn to predict multiple event cues that can be used at test-time to \emph{choose} which task to solve. Prior work has aimed to achieve similar results by conditioning the policy on a goal or objective indicator~\cite{schaul2015universal,Codevilla2018}. Our approach does not need to choose specific goals to solve during training -- all learning is off-policy, and an explicit goal representation is not provided, only a set of event cues that might be relevant for a test-time task. Our approach is similar in spirit to Horde~\cite{sutton2011horde}, which also learns to predict future occurrences of specific sensory readings. Unlike our work, this prior method does not demonstrate control for new compositions of learned events. \citet{dosovitskiy2016learning} demonstrate composition of multiple predictors, but condition on a single action without bootstrapping, and therefore requires on-policy training. By conditioning on a sequence of actions, we obtain unbiased policies with off-policy training.

In addition to proposing a composable reinforcement learning framework based on prediction, our work makes a conceptual contribution in how computer vision systems can be used within a learning-based control framework: we propose that, for commonly attainable visual recognition signals (such as object detections), reinforcement learning can effectively convert these \emph{detections} into \emph{predictions} by using them in place of reward signals in a multi-event prediction framework. Prior work has incorporated detectors into reinforcement learning-based settings~\cite{Devin2018_ICRA}, but for training a model-free policy that is only designed to generalize to new instances of the same event cues. Prediction has been an active area of research in computer vision, including prediction of future images~\citep{mathieu2016deep,finn2016unsupervised,denton2017unsupervised}, motion~\citep{vondrick2015anticipating}, and even physical events~\citep{mottaghi2016happens,pinto2016supersizing,Kahn2018_ICRA}. We study action-conditioned prediction, which enables us to control a robot to bring about desired combinations of predicted events at test-time.

There has been extensive work on autonomous robot navigation~\citep{rosen1968application,thorpe1988vision,urmson2008autonomous}. Much of the prior work on learning-based robot navigation has focused on imitation learning~\cite{pomerleau1989alvinn,muller2006off,ross2011reduction,wulfmeier2016watch,bojarski2016end}, which requires demonstrations. Our method does not assume access to demonstrations, and can learn from off-policy data. Supervised learning for navigation using off-policy data has been investigated, including learning drivable routes~\cite{barnes2017find} and near-to-far obstacle detectors~\cite{hadsell2009learning}. Our approach is similar in that we also predict future events, but prior works typically rely on a hand-engineered control policy, while our control policy using CAPs can improve with more data because it is conditioned on the robot's intended actions. RL approaches~\cite{SuttonBarto} are designed to learn and improve control policies from data, including from off-policy data~\cite{Watkins1992_ML}.  Much of the reinforcement learning work for robot navigation has focused on collision avoidance~\citep{Richter2017_RSS,Chen2017_IROS,Kahn2018_ICRA}, while our work addresses goal-directed, multi-objective navigation. RL-based approaches for learning goal-directed navigation have been proposed~\cite{mirowski2016learning,sadeghi2017cad}, but typically learn using hand-crafted reward signals that are difficult to calculate in the real-world. In contrast, our approach directly addresses the issue of defining the reward signals in the real-world by turning deployable vision-based detectors into predictors using our CAPs.


\section{Composable Action-Conditioned Predictors (CAPs)}


We now present our composable action-conditioned predictors (CAPs) framework. CAPs learns to ``back up'' a set of event cues into the past using a predictive model. This model takes as input the state and a sequence of planned future actions, and outputs predictions of these event cues, which can consist of anything relevant to the robot's task, such as collision, speed, and road lane positions. Using this model, the user can then specify the reward function they wish the robot to maximize in terms of the event cues.


\subsection{The CAPs Model}
\label{sec:caps-model}


Formally, CAPs correspond to a model $f_\theta(\bs_t, \bA_t^H) \rightarrow \hat{E}_t^{(H,I)}$, which is a function parameterized by parameters $\theta$ that maps the state $\bs_t$ at time $t$ and a sequence of $H$ intended actions ${\bA_t^H = (\ba_t, ..., \ba_{t+H-1})}$ to predicted future event cues $\hat{E}_t^{(H,I)}$. The event cues $\hat{E}_t^{(H,I)}$ consist of events $\hat{e}_{t+h}^{(i)}$, where $h \in \{0, ..., H-1\}$ indexes the prediction time length and $i \in \{0, ..., I-1\}$ indexes the $i$th event cue. This model can be viewed as an extension of \citet{Kahn2018_ICRA} to multiple event cues. The model is trained on a dataset of state-action-event tuples $\mathcal{D} := \{\bs_t, \bA_t^H, E_t^{(H,I)}\}$ such that
\vspace*{-5pt}
\begin{align}
\theta^* &= \arg\min_{\theta} \sum_{(\bs_t, \bA_t^H, E_t^{(H,I)}) \in \mathcal{D}} \sum_{h} \sum_{i} \| \hat{e}_{t+h}^{(i)} - e_{t+h}^{(i)} \|.~\label{eqn:train-pred}
\end{align}
An important distinction between our approach and that of \citep{dosovitskiy2016learning} is that our model is conditioned on a sequence of actions, which is critical for off-policy training. \citep{dosovitskiy2016learning} predicts events far in the future conditioned on the single current action, which implicitly makes the predictor conditional on a policy, necessitating on-policy data collection. Training the CAPs is completely off-policy: all data collected by the robot can be used for training, which is advantageous for real-world robot learning in which gathering data is laborious and expensive.


\subsection{Autonomous Labeling of Event Cues}
\label{sec:caps-labelling}

Although the training dataset could be generated by letting the robot act in the environment and hand-labeling the event cues $E_t^{(H,I)}$, the amount of human supervision needed would become prohibitively expensive. We instead opt for automated labeling by leveraging existing detection systems to label these event cues. These detection systems, including modern computer vision systems, enable our approach to predict cues about the environment that would have otherwise remained unknown. Consequently, by having access to these labeled event cues and training our CAPs to predict these event cues, CAPs can be used to achieve tasks that would have otherwise required \textit{a priori} knowledge of the environment. In short, the detection system provides ``what'' the robot wants to learn about by labelling the event cues, while CAPs provide ``how'' the robot can take actions to achieve these events. We discuss specific event cues and their respective learned detection systems used for labeling in the context of autonomous robot navigation in Section~\ref{sec:caps-robonav}.

In addition to leveraging existing detection systems, we also can label a subset of the cues using self-supervision. For example, if we want the robot to move at a desired speed, and its state includes speed, it can self-label. Although few event cues can be self-supervised, these self-supervised signals are robust.


\subsection{Action Selection}
\label{sec:caps-reward}

Using the trained CAPs model $f_\theta$ and following the approach of~\citep{Kahn2018_ICRA}, the user can encode a task that the robot must accomplish by defining a reward function $R(\hat{E}_t^{(H, I)})$ on the predicted future event cues. The robot can then calculate the action sequence that maximizes this reward function by solving the following optimization:
\begin{align}
\bA_t^* = \arg\max_{\bA^H} R(\hat{E}_t^{(H, I)}) : \hat{E}_t^{(H, I)} = f_\theta(\bs_t, \bA^H). \label{eqn:opt}
\end{align}
We use stochastic optimization based on the cross entropy method~\citep{Rubinstein1999CEM} to solve this maximization, although in principle any optimizer could be used.

Although the procedure in Eqn.~\ref{eqn:opt} enables the robot to plan $H$ time steps into the future, we would like our robot to be able to accomplish tasks beyond this horizon. We therefore adopt a model predictive control (MPC) approach, in which the robot solves for the optimal action sequence at each time step using Eqn.~\ref{eqn:opt}, executes the first action, proceeds to the next state, and repeats the planning procedure. Although this MPC approach is not equivalent to planning for the entire horizon of the task, it has been shown to be an effective and robust method for robot control~\cite{camacho2013model}.


\subsection{Off-Policy Learning and Policy Deployment}

We now describe the full off-policy learning algorithm using CAPs. A dataset of state-action pairs $(\bs_t, \ba_t)$ is gathered by having the robot act in the environment according to some policy, such as a random exploration policy or the latest trained CAPs policy (Eqn.~\ref{eqn:opt}). The robot does not need data gathered for the tasks it will need to accomplish at test time, but it does require data in which the relevant event cues are present. The dataset is then used by the detection systems to label the event cues $(e_t^{(0)}, ..., e_t^{(I-1)})$, which are then added back into the dataset. Finally, the CAPs model is trained using this dataset (Eqn.~\ref{eqn:train-pred}).
After running off-policy learning, the user can encode a desired task for the robot to accomplish (Sec.~\ref{sec:caps-reward}). Importantly, the user can change the task without having to re-train CAPs, so long as the task is expressed in terms of the event cues that CAPs has learned to predict.


\begin{figure}[b]
	\vspace*{-15pt}
    \centering
    \includegraphics[width=\textwidth]{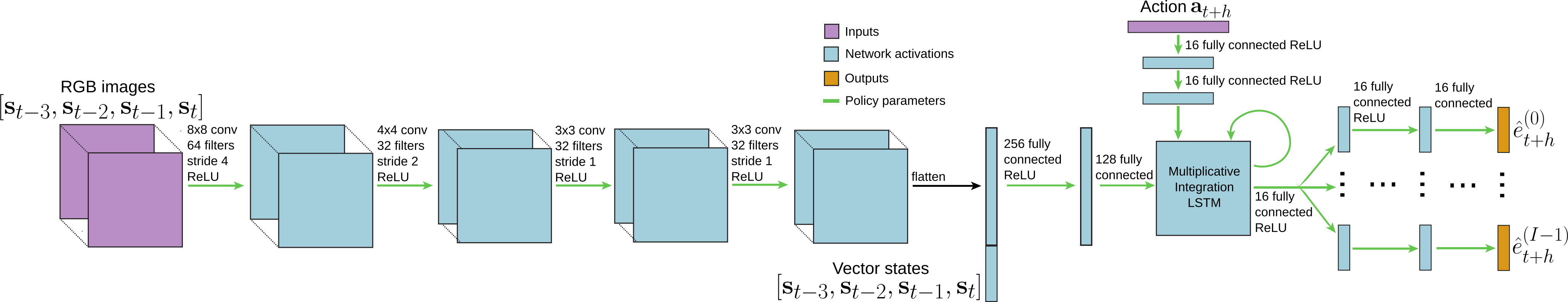}
    \caption{The composable action-conditioned predictions (CAPs) network architecture. The network first processes the past four RGB images with convolutional layers, concatenates the result with the past four vector states, and then passes the concatenation through fully connected layers to form the initial hidden state of the recurrent neural newtork (RNN). The RNN, which is a multiplicative integration LSTM~\cite{wu2016multiplicative}, sequentially processes each action, and the resulting hidden layer is used to predict each event cue.}
    \label{fig:nav-comp-graph}
\end{figure}

\section{CAPs for Robot Navigation}
\label{sec:caps-robonav}
We now instantiate our CAPs algorithm for robot navigation. Robot navigation is an ideal testbed for CAPs because navigation is a multi-objective task in which many of the event cues---such as road lane and object locations---are unknown \textit{a priori}. To instantiate CAPs, we must instantiate the CAPs model (Sec.~\ref{sec:caps-model}), event cue labellers (Sec.~\ref{sec:caps-labelling}), and the task reward function (Sec.~\ref{sec:caps-reward}).

\textbf{The CAPs model.} We instantiate the CAPs model by defining its inputs, outputs, and parameterization. The inputs to the CAPs model are state observations $\bs$ and actions $\ba$. The state observations include all readings from sensors on board the robot, such as cameras, wheel encoders, collision bumpers, and inertial measurements. For a ground robot, the actions consist of the desired steering angle and speed. The outputs of the predictive model consist of event cues relevant to goal-oriented navigation, such as road lanes, collisions, speeds, and headings.

We parameterize the model using a deep neural network in order to make it feasible to input high-dimensional observations, such as images. This model is depicted in Fig.~\ref{fig:nav-comp-graph}. The network first passes the input images through convolutional layers, concatenating the result with the input state vectors, and processes the concatenation through additional fully connected layers. The final layer serves as the initial hidden state for a recurrent network, which sequentially processes each planned action. The event cues at each time step are then predicted by processing the RNN hidden state with separate fully connected layers.

\textbf{Event cue labeling.}  For the vision-based event cues, such as road lanes and doorways, we use modern computer vision models. Specifically, we train FCNs~\citep{Long2015_CVPR} to segment relevant event cues from the onboard camera images. The data used to train these models is a subset of the data used to train the CAPs model, and was either labeled by the simulator for simulation experiments, or by a human for real-world experiments. The non-vision-based event cues are simple functions of the robot state, such as if the robot collided or not, and are extracted automatically.

\textbf{Task reward function.} We encode desired tasks as linear combinations of reward functions of each individual event cue: $R(\hat{E}_t^{(H,I)}) = \sum_{t'=t}^{t+H-1} \sum_{i} \alpha^{(i)} \cdot R^{(i)}(\hat{e}_{t'}^{(i)})$, where $\alpha^{(i)}$ indicates the relative importance of each reward function. For example, the collision reward function will be highly weighted to ensure the robot does not collide. With this formulation, the task we want the robot to accomplish can be specified and altered by setting the weights $\alpha^{(i)}$.


\section{Experiments}

\newcommand{\Qflexible}{\textbf{Q1}}
\newcommand{\Qoffpolicy}{\textbf{Q2}}
\newcommand{\Qvision}{\textbf{Q3}}

We now present results evaluating our approach on simulated and real-world ground robot navigation tasks. In our evaluation, we aim to answer the following questions:
\begin{itemize}
	\vspace*{-8pt}
    \item[\Qflexible] Does our event cue prediction approach enable flexible behavior at test time?
    \item[\Qoffpolicy] Is our approach able to learn from off-policy data?
    \item[\Qvision] Are we able to learn event predictors using learned detection models, such as modern computer vision systems?
   	\vspace*{-8pt}
\end{itemize}

In evaluating CAPs, we compare with prior methods that can scale to image observations, learn from off-policy data, accomplish various tasks at test time, and are sample-efficient. We therefore compare with goal-conditioned deep Q-learning (GC-DQL)~\cite{sutton2011horde,schaul2015universal,Andrychowicz2017_NIPS}, in which a neural network Q-function is learned that also takes as input the desired goal. We also compared with goal-conditioned deep Q-learning in which each subreward is learned separately (GC-DQL-sep), which is similar to~\citet{Haarnoja2018_ICRA}, but without entropy maximization. We do not compare with model-based methods because they either assume access to the ground-truth robot and environment state, or require large amounts of data to learn the dynamics of raw image observations.

We consider three experiments: a simulated path following task in a forest, a simulated goal-directed navigation task in a city environment, and a real-world indoor navigation task with an RC car. For each task, in order to instantiate CAPs, we must define which event cues $\hat{e}^{(i)}$ the model will be predicting, how the event cue labels $e^{(i)}$ will be generated, and what is the task reward function $R(\hat{E}_t^{(H,I)})$. Code, videos, and additional details are available at \rurl{github.com/gkahn13/CAPs}.

\begin{figure}[b]
    \centering
    \begin{subfigure}[c]{0.19\textwidth}
        \caption*{GC-DQL}
        \includegraphics[height=0.12\textheight]{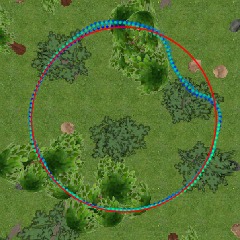}
        \includegraphics[height=0.12\textheight]{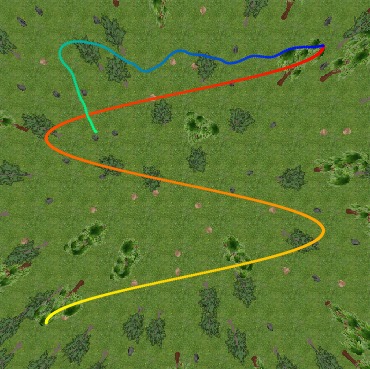}
        \label{fig:exp-cpf-dql}
    \end{subfigure}
    \hfill
    \begin{subfigure}[c]{0.19\textwidth}
        \caption*{GC-DQL-sep}
        \includegraphics[height=0.12\textheight]{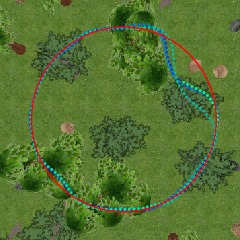}
        \includegraphics[height=0.12\textheight]{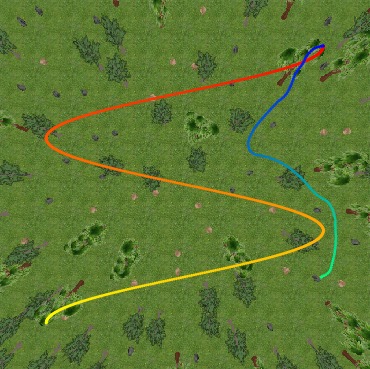}
        \label{fig:exp-cpf-bootstrap}
    \end{subfigure}
    \hfill
    \begin{subfigure}[c]{0.19\textwidth}
        \caption*{CAPs (ours)}
        \includegraphics[height=0.12\textheight]{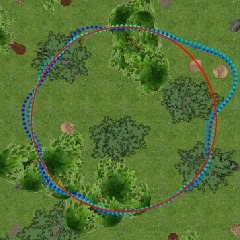}
        \includegraphics[height=0.12\textheight]{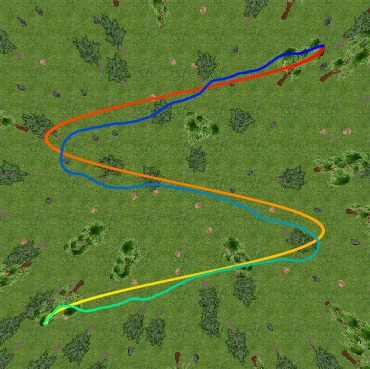}
        \label{fig:exp-cpf-ours}
    \end{subfigure}
    \hfill
	\begin{subfigure}[c]{0.4\textwidth}
		\centering
		\begin{tabular}{|c|c|c|}
	        \hline             & \specialcell{Avg \%\\Success}  & \specialcell{Avg Dist.\\From Path} \\
	        \hline GC-DQL      &  0.0\%          & 5.01 $\pm$ 3.76      \\
	        \hline GC-DQL-sep  &  8.3\%          & 6.03 $\pm$ 3.35      \\
	        \hline CAPs (ours) & \textbf{75.0\%} & \textbf{2.02 $\pm$ 1.60}      \\
	        \hline
	    \end{tabular}
	    \caption{Evaluation on novel sinusoidal paths.}
	    \label{tab:cpf}
	\end{subfigure}
    \caption{Comparison of our CAPs approach against goal-conditioned deep Q-learning (GC-DQL) and separated goal-conditioned deep Q-learning (GC-DQL-sep) on the task of path following. Each approach was trained via reinforcement learning to following a circular path and avoid collisions (top row images), which shows a birds-eye view of the desired trajectory in red hues and actual trajectory in blue hues. Each approach was then evaluated on a sinusoidal path (bottom row images). Although all approaches successfully learned on the training task, only our approach (Table~\ref{tab:cpf}) was able to generalize to a different task at test-time.}
    \label{fig:exp-cpf}
\end{figure}

\textbf{Simulated forest.} The first task is a path following task in a forest-like environment (Fig.~\ref{fig:exp-cpf}) built on the Bullet physics engine~\citep{coumans2013bullet} with Panda3d~\citep{goslin2004panda3d} for rendering. The goal is to stay close to the path while avoiding collisions.

We define two event cues: the probability of collision and heading. The labels come from self-supervision, because collision and heading can be labelled directly from the robot's sensor observations. The reward function at test-time is:
\begin{align*}
R(\hat{E}_t^{(H,I)}) = \sum_{t'=t}^{t+H-1} 500 \cdot (1 - \hat{e}_{t'}^{(coll)}) + (\cos(\hat{e}_{t'}^{(heading)} - \textsc{goal\_heading}) - 1).
\end{align*}
In order to gather data, we let our approach and the prior works each run RL on the task of following a circular path. Fig.~\ref{fig:exp-cpf} shows that all approaches successfully learn to follow the circular path without colliding. We chose to gather data in this way because Q-learning approaches, while technically off-policy, are still sensitive to the data distributions when using neural networks. We therefore wanted to give the prior methods the best chance of success by giving them access to data that was partially on-policy.

\begin{figure}[!b]
    \centering
    \begin{subfigure}[b]{\textwidth}
	    \begin{subfigure}[b]{0.32\textwidth}
    	    \caption*{GC-DQL}
	        \includegraphics[width=\textwidth]{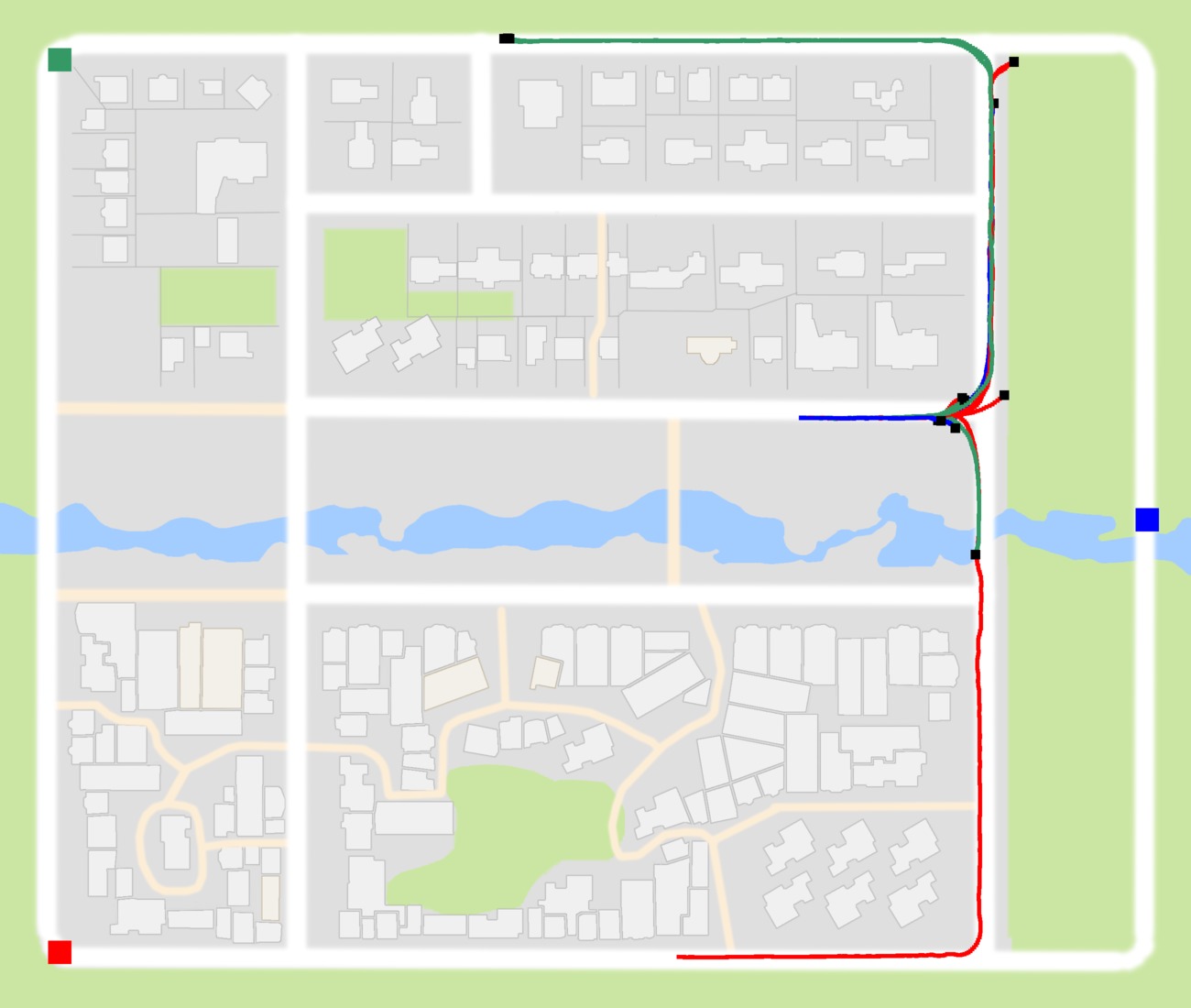}
	        \label{fig:exp-carla-map-dql}
	    \end{subfigure}
	    \hfill
	    \begin{subfigure}[b]{0.32\textwidth}
	        \caption*{GC-DQL-sep}
    	    \includegraphics[width=\textwidth]{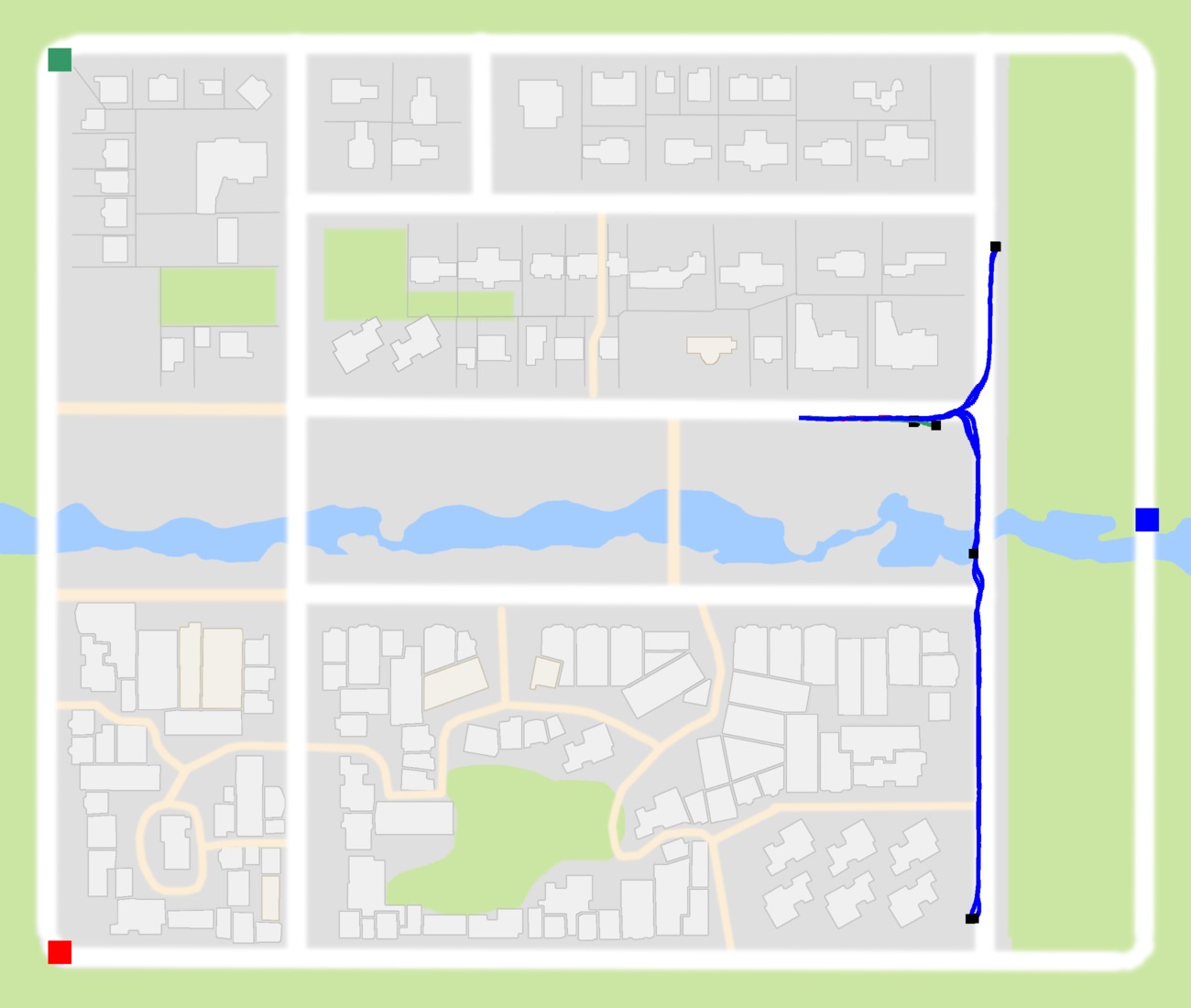}
	        \label{fig:exp-carla-map-dql-sep}
	    \end{subfigure}
	    \hfill
	    \begin{subfigure}[b]{0.32\textwidth}
	        \caption*{CAPs (ours)}
	        \includegraphics[width=\textwidth]{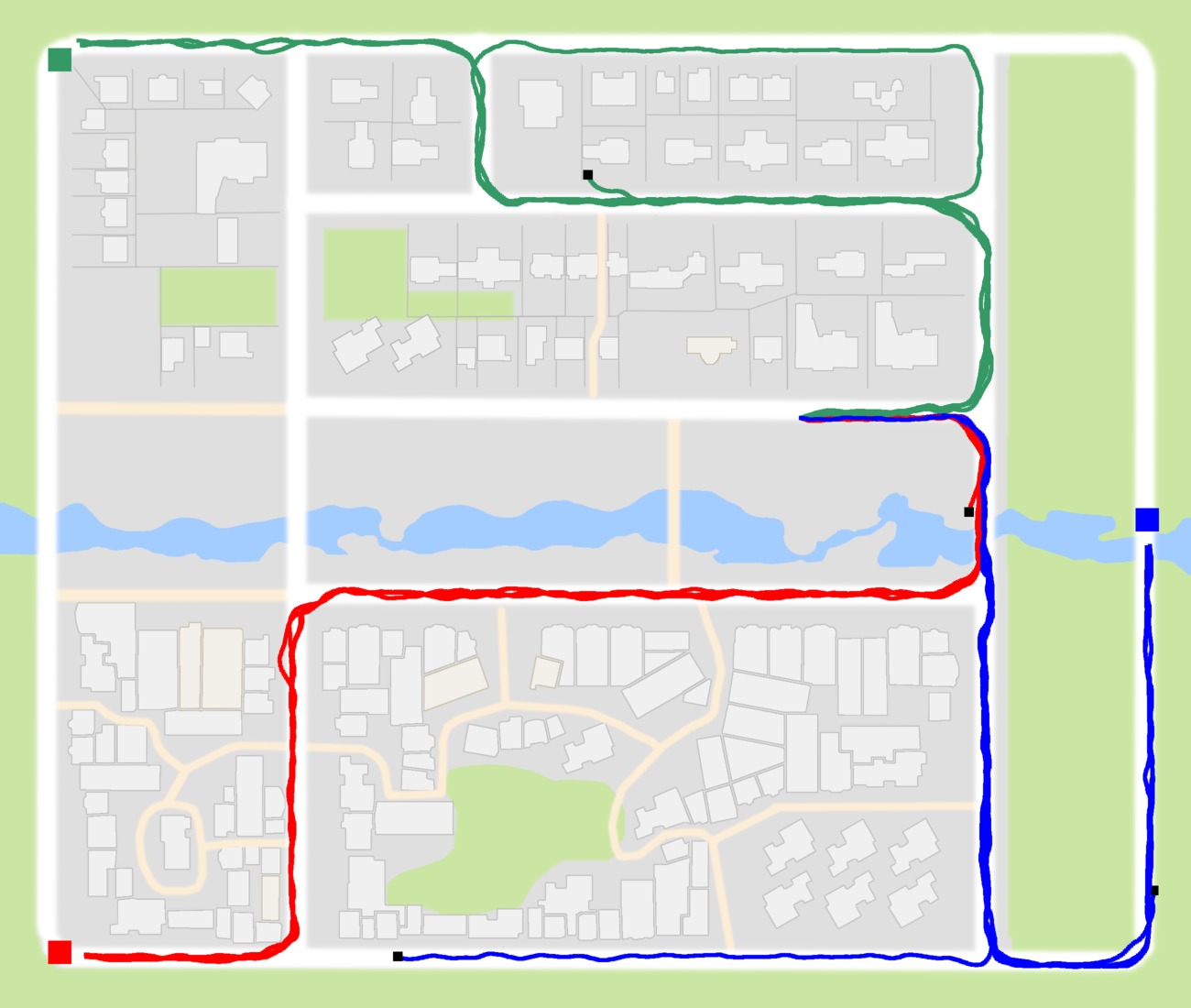}
	        \label{fig:exp-carla-map-ours}
	    \end{subfigure}
		\vspace{-10pt}
	    \caption{Birds-eye view for each approach of the resulting trajectories when attempting to go to either the red, green, or blue goal location (five attempts per goal location). Our CAPs approach is able to reach the goal the majority of the time, while the prior methods crash before reaching the goal.}
	    \label{fig:exp-carla-map}
	\end{subfigure}
	
    \begin{subfigure}[c]{0.49\textwidth}
		\centering
		{\small
		\begin{tabular}{|c|c|c|c|}
		\hline & GC-DQL & GC-DQL-sep & CAPs (ours) \\
		\hline \specialcell{\% did\\not crash} & 7 & 0 & \textbf{73} \\
		\hline \specialcell{\% reached\\goal} & 0 & 0 & \textbf{67} \\
		\hline \specialcell{speed\\(m/s)} & 3.6 (1.3) & 5.6 (0.2) & \textbf{6.7 (0.2)} \\
		\hline \specialcell{\% in\\right lane} & 98 (3) & \textbf{100 (6)} & 74 (8) \\
		\hline
		\end{tabular}				
		}
		\caption{Corresponding quantitative results for (a).}
		\label{table:exp-carla}
	\end{subfigure}%
	\hfill
	\begin{subfigure}[c]{0.49\textwidth}
		\includegraphics[width=0.32\textwidth]{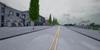}
		\hfill
		\includegraphics[width=0.32\textwidth]{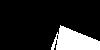}
		\hfill
		\includegraphics[width=0.32\textwidth]{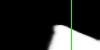}
		\caption{To enable our CAPs approach to learn to stay in the right lane, we trained a model to take as input the onboard RGB image (left) and predict where the road is (right) using labels (center). Once trained the learned segmentation model autonomously labels where the center of the right lane is (green vertical line), the CAPs model can learn to keep the center of the right lane in the middle of its camera view.}
		\label{fig:exp-carla-learnedroad}
	\end{subfigure}	
	\caption{Comparison of our CAPs approach against prior methods on the task of reaching a target goal position while avoiding collisions, driving at 7m/s, and staying in the right lane in the CARLA simulator~\cite{Dosovitskiy2017_CoRL}.}
	\label{fig:exp-carla}
\end{figure}

After training, we evaluated each approach on following sinusoidal paths in different parts of the forest environment. Note that these paths never occurred in the training data. Fig.~\ref{fig:exp-cpf} shows qualitative results for each method, while Table~\ref{tab:cpf} provides a quantitative comparison. Our approach avoids collisions and closely follows the sinusoidal path, only deviating when obstacles are on the path itself, while the prior methods crash often and do not closely follow the path. This result shows that even though the prior methods were successful on the circular path they trained on during reinforcement learning, they fail to generalize at test time to different tasks. In contrast, CAPs can flexibly accomplish different tasks at test time (\Qflexible).

\textbf{Simulated city.} The second task we evaluate our approach on is goal-directed navigation in a city environment using CARLA~\cite{Dosovitskiy2017_CoRL}, a driving simulator with realistic renderings and physics.
The objective is to reach the goal while avoiding collisions, driving at a desired speed, and staying in the lane. We train CAPs with five event cues: probability of a collision, heading, speed, is the right lane visible, and the pixel distance of the center of the right lane to the center of the camera image. The labels for the collision, heading, and speed event cues come from self-supervision because these values come directly from the robot's sensor observations. However, we use a learned computer vision system---specifically, a segmentation model~\cite{Long2015_CVPR}---to label the lane event cues; Fig.~\ref{fig:exp-carla-learnedroad} shows an example input image, the ground truth segmentation label, and the model's predicted segmentation and resulting prediction of the center of the right lane. The reward function at test-time is:

{\small
\vspace*{-17pt}
\begin{align*}
R(\hat{E}_t^{(H,I)}) = \sum_{t'=t}^{t+H-1} &50 \cdot (1 - \hat{e}_{t'}^{(coll)}) - 
                                     3 \cdot \frac{| \hat{e}_{t'}^{(speed)} - \textsc{goal\_speed} |}{\textsc{goal\_speed}} + 
                                     5 \cdot \hat{e}_{t'}^{(lane\_seen)} (1 - | \hat{e}_{t'}^{(lane\_diff)} | ) \\
                                     &- \frac{5}{\pi} \cdot | \hat{e}_{t'}^{(heading)} - \textsc{goal\_heading} | -
                                     0.15 \cdot \| \ba_{t'}^{(steer)} \|_2^2.
\end{align*}
\vspace*{-15pt}
} 

For training data, we ran DQL for the task of collision avoidance, resulting in 800,000 points (2.3 days worth) of data. We chose to gather data using DQL to give the prior Q-learning-based methods the best chance at success, and to demonstrate that CAPs can learn from off-policy data (\Qoffpolicy). We then trained all methods on this data for the combined task of collision avoidance, driving at a desired speed, and staying in the right lane.

We evaluated all approaches on driving at 7 m/s and trying to reach one of three destination locations. Fig.~\ref{fig:exp-carla-map} shows a birds-eye view of the city map and the resulting trajectories of each method's policy, while Table~\ref{table:exp-carla} provides the corresponding quantitative results. Our CAPs policy is able to successfully reach each goal the majority of the time while driving in the right lane at the desired speed, while the prior methods are never able to reach the goal.

The key aspect of CAPs that enables its success is its flexibility at test time (\Qflexible): we can train the CAPs model once using off-policy data, and then define reward function at test time to achieve the desired robot behavior for the considered task. Although defining this reward function is non-trivial, it is significantly easier and less time consuming than designing the reward function for standard RL algorithms. For example, we spent only two hours tuning the CAPs reward function for this task, while running GC-DQL or GC-DQL-sep just once takes 12 hours; considering that we tried dozens of reward functions during the two hours of tuning with CAPs, attempting to do this amount of tuning with a standard RL algorithm could take days or weeks, and possibly still not result in a successful policy.

Another key aspect of our approach is using learned detection models to autonomously label the event cues (\Qvision). Although using a learned segmentation model (Fig.~\ref{fig:exp-carla-learnedroad}) in simulation is not strictly necessary because we have access to the ground truth labels, learned event cue labellers will be crucial in real-world experiments in which ground truth labels from humans is prohibitively expensive to obtain. We further demonstrate the importance of autonomous labelling in the following real-world experiments.

\begin{figure}[!b]
    \centering
    \begin{subfigure}[b]{\textwidth}
        \includegraphics[width=0.19\textwidth]{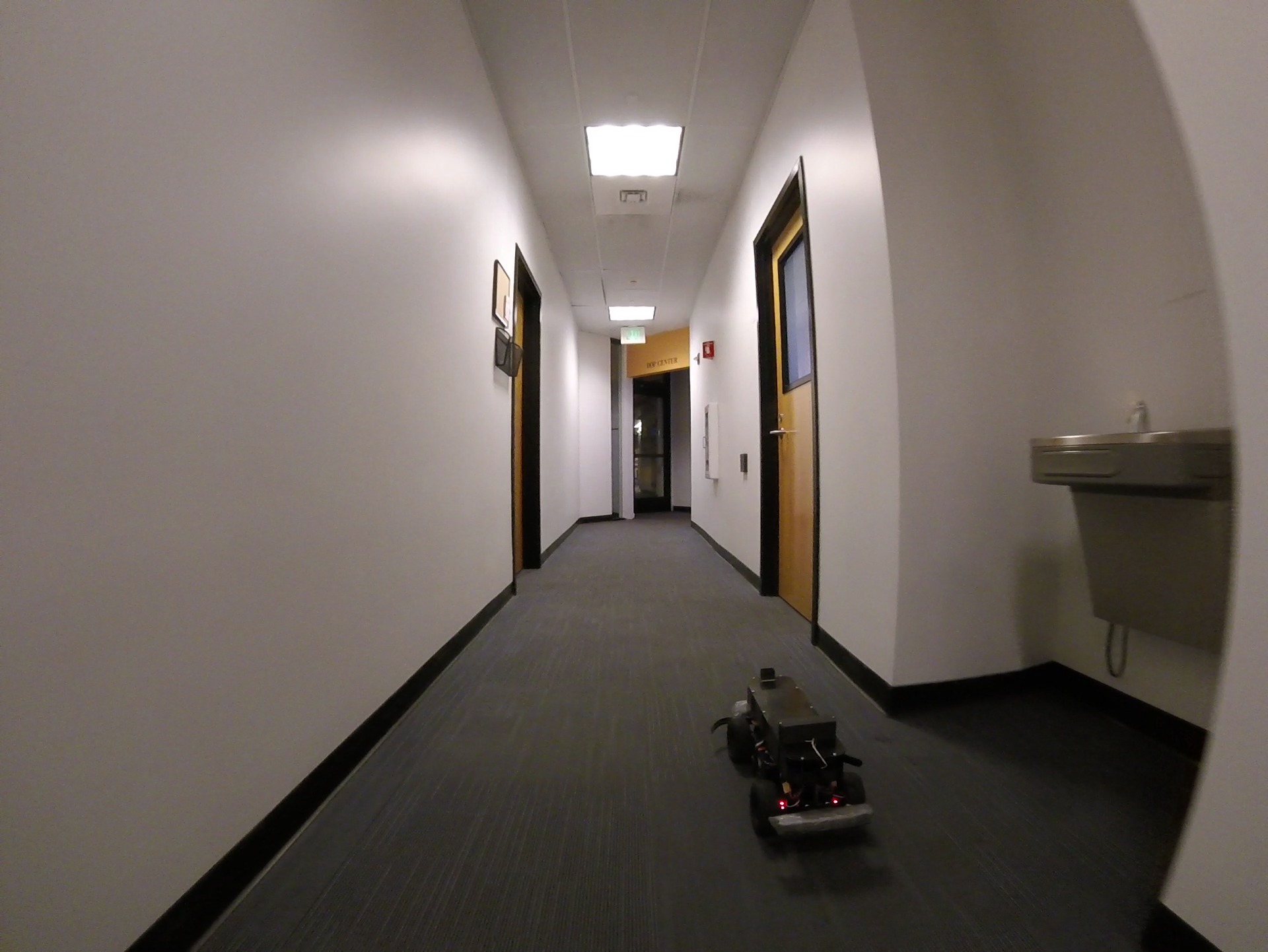}
        \hfill
        \includegraphics[width=0.19\textwidth]{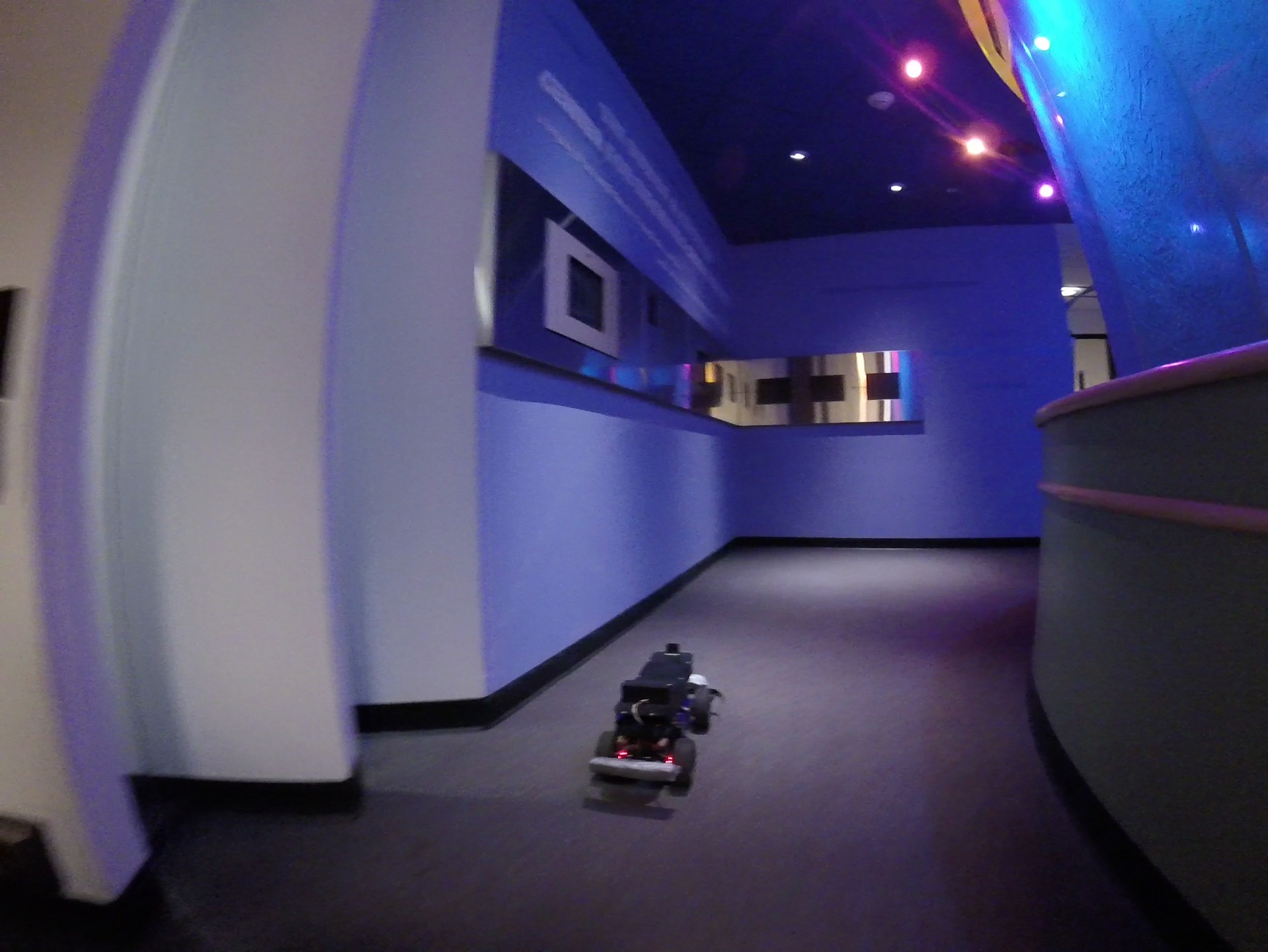}
        \hfill
        \includegraphics[width=0.19\textwidth]{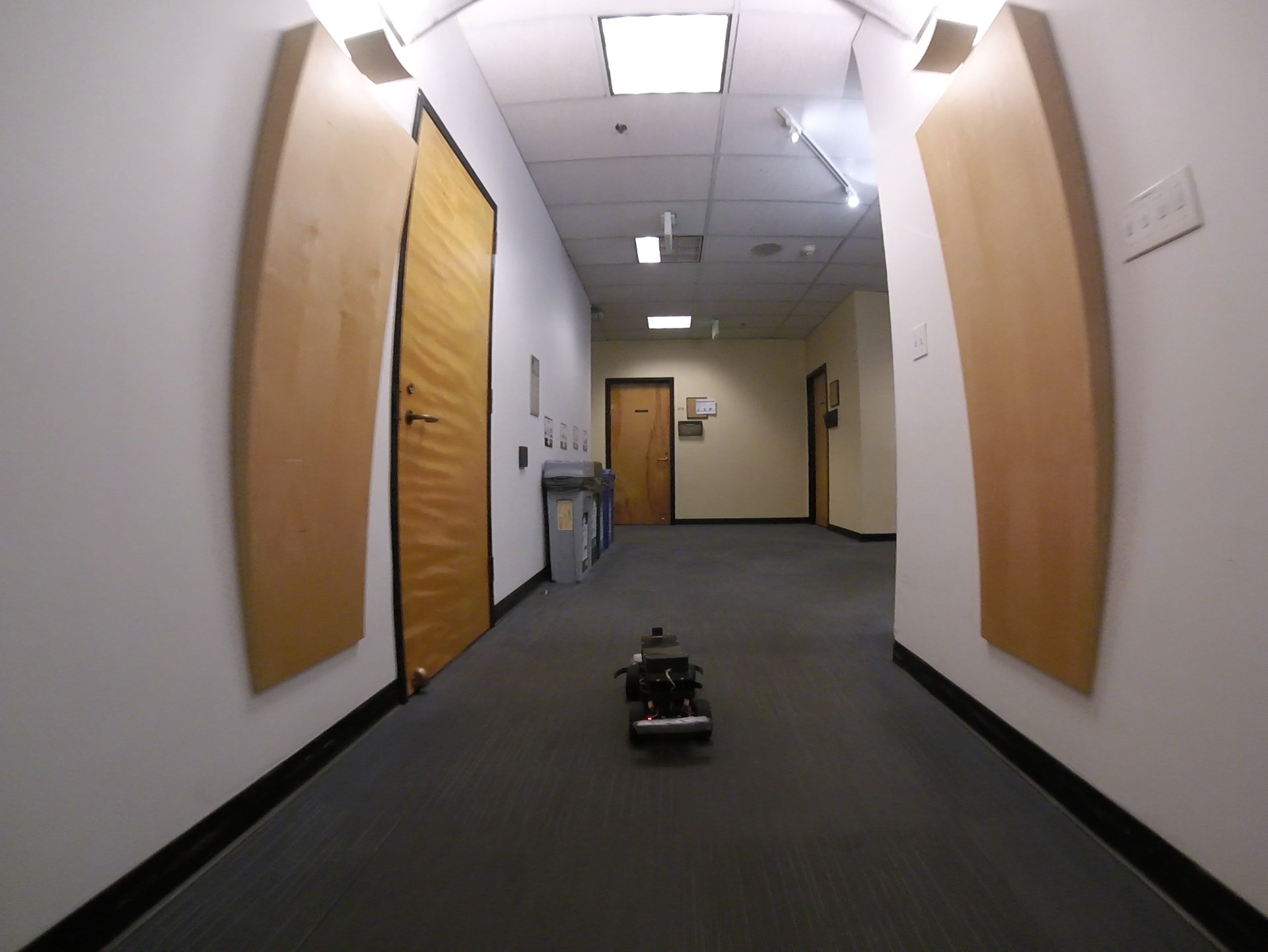}
        \hfill
        \includegraphics[width=0.19\textwidth]{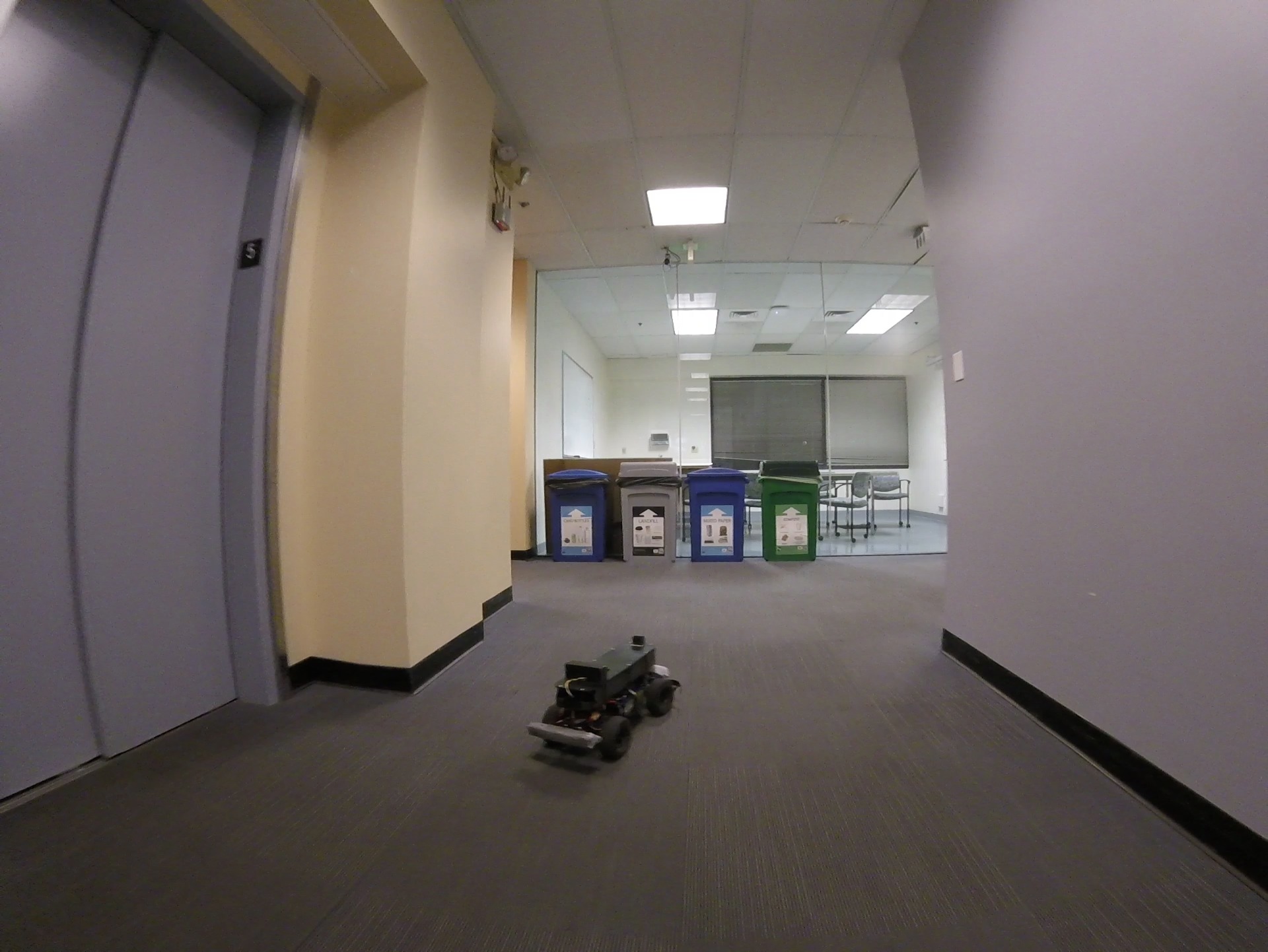}
        \hfill
        \includegraphics[width=0.19\textwidth]{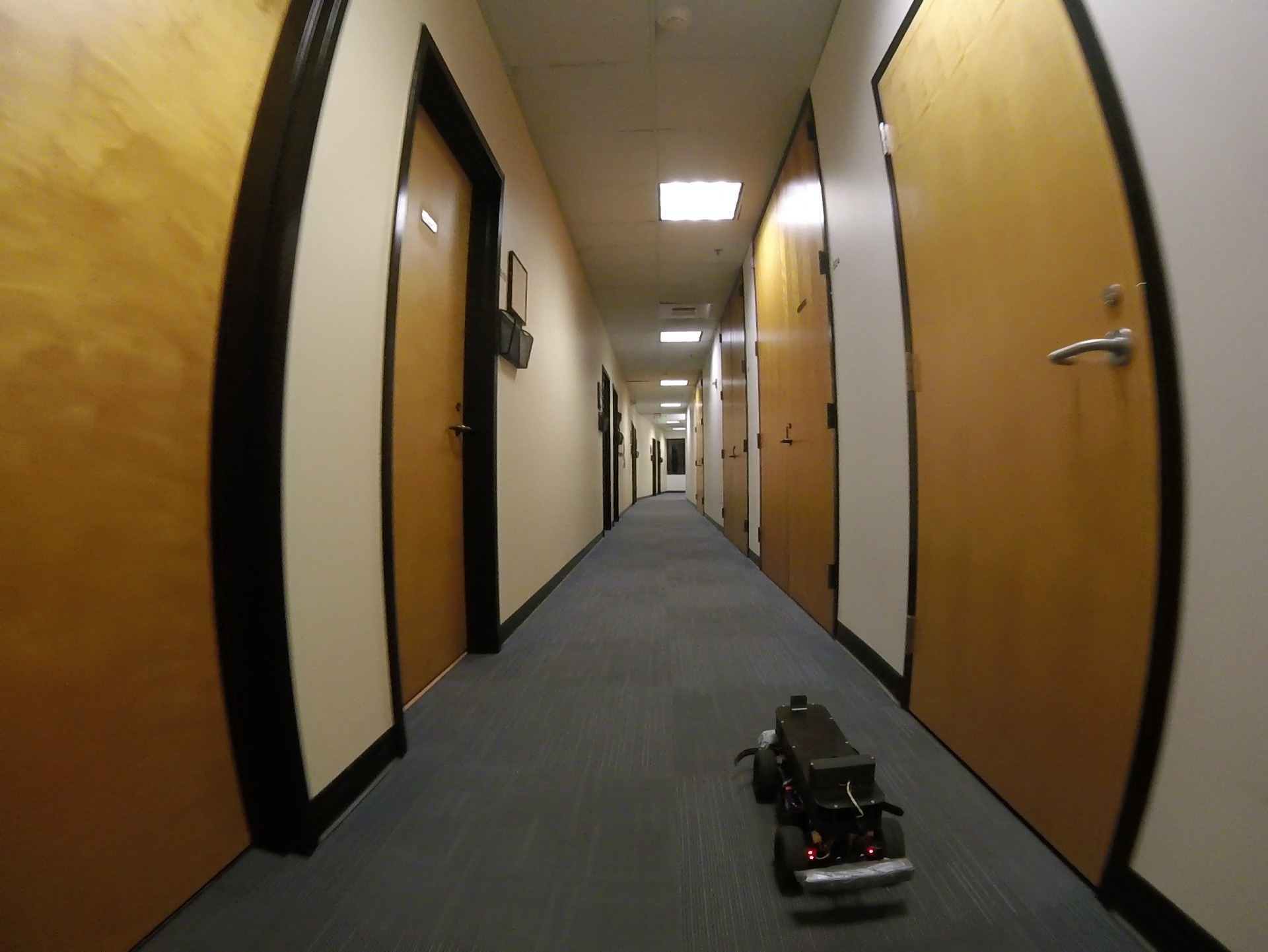}
    \end{subfigure}
    \begin{subfigure}[b]{\textwidth}
        \includegraphics[width=0.19\textwidth]{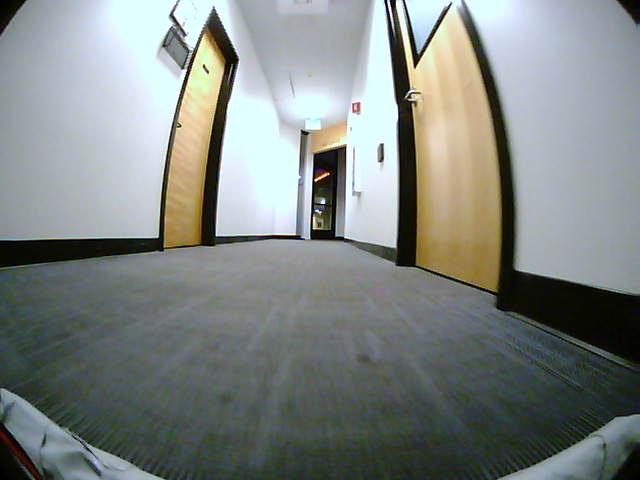}
        \hfill
        \includegraphics[width=0.19\textwidth]{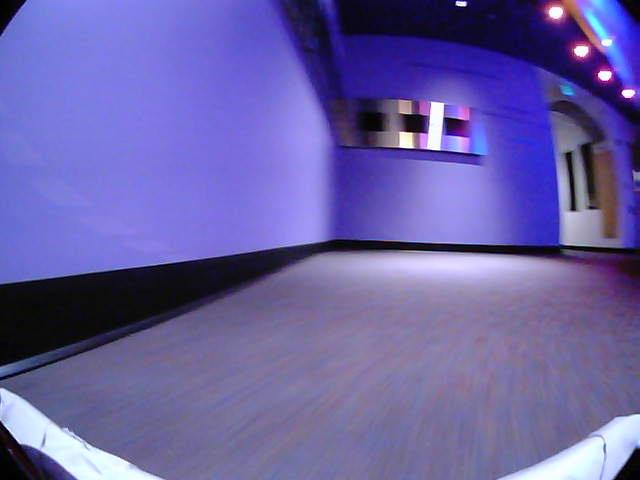}
        \hfill
        \includegraphics[width=0.19\textwidth]{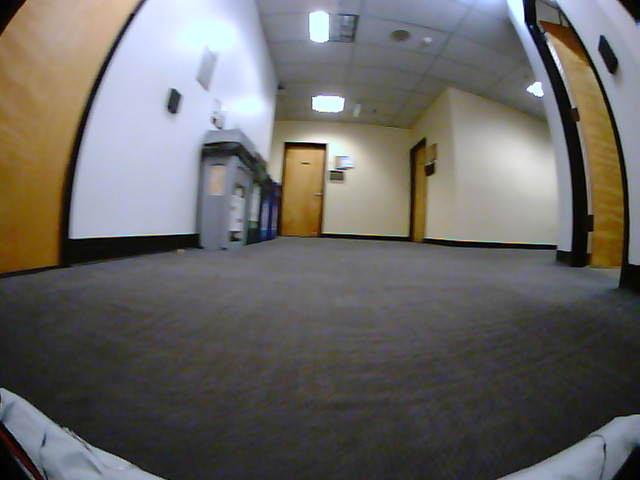}
        \hfill
        \includegraphics[width=0.19\textwidth]{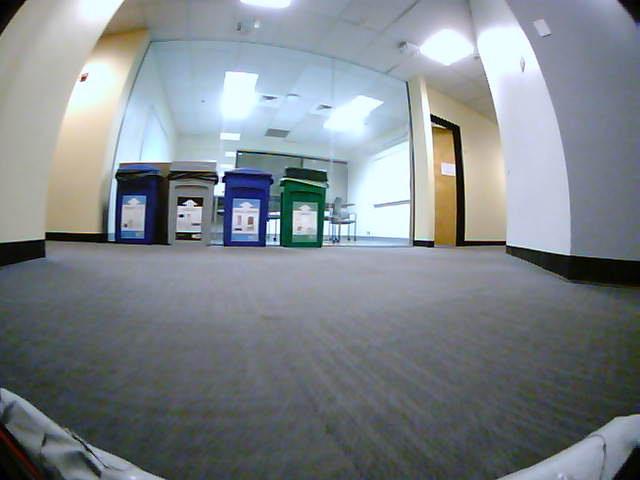}
        \hfill
        \includegraphics[width=0.19\textwidth]{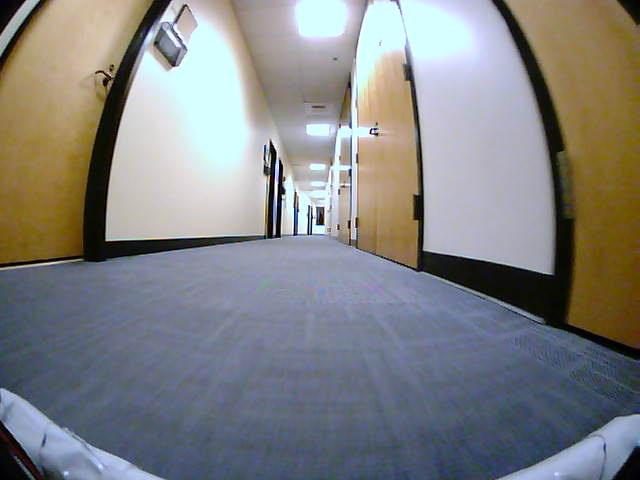}
    \end{subfigure}
    \caption{Third-person images (top row) and first-person images from the onboard camera (bottom row) from the RC car navigating through the 5th floor of Cory Hall at UC Berkeley for the task of avoiding collisions, following desired goal headings, and going towards doors using our CAPs approach.}
    \label{fig:exp-rccar-tpv-fpv}
\end{figure}

\textbf{Real-world indoor navigation.} The third task we evaluate our approach on is a real-world goal-directed indoor navigation task with an RC car (Fig.~\ref{fig:rccar-teaser}).
The task objective is to avoid collisions, follow the desired goal headings, and go near doors, which simulates a package delivery application. We therefore define three event cues: probability of a collision, heading, and the fraction of the image that is a door. The labels for the collision and heading come directly from the robot's sensor observations, while we use a learned computer vision segmentation model to label the door fraction. We trained the segmentation model by labelling 345 (0.2\%) of the images from the RL-gathered dataset. The reward function is:

{\scriptsize
\vspace*{-10pt}
\begin{align*}
R(\hat{E}_t^{(H,I)}) = \sum_{t'=t}^{t+H-1} (1 - \hat{e}_{t'}^{(coll)}) \cdot \left[ 1 - \frac{0.1}{\pi} \cdot | \hat{e}_{t'}^{(heading)} - \textsc{goal\_heading} | + 0.05 \cdot \hat{e}_{t'}^{(door\_frac)} \right] - 0.01 \cdot \|\ba_{t'}\|_2^2.
\end{align*}
\vspace*{-10pt}
}

For training data, we ran RL with CAPs on the 5th floor of Cory Hall at UC Berkeley for 11 hours, gathering 158,400 data points. The RL objective for gathering this data was purely collision avoidance, and therefore the data is off-policy (\Qoffpolicy). We then evaluated all approaches in this same environment. Although this evaluation does not test our learned policy's ability to generalize to new environments, our approach does not need to rely on policy generalization; instead, we rely on our sample-efficient and flexible real-world CAPs learning algorithm to quickly learn in new environments.

\begin{wraptable}{r}{0.7\textwidth}
	\vspace*{-12pt}
	\centering
	{\small
	\begin{tabular}{|c|c||c|c|c|c|}
		\hline Experiment & Method & \specialcell{\% of Route\\Traversed} & \specialcell{\% Did\\Not Crash} & \specialcell{\% Successful\\Turns} & \specialcell{Avg. Door\\\% Seen} \\
		
		\hline \hline \multirow{2}{*}{Collision} & DQL & 100 (0) & 100 & N/A & N/A \\
		\cline{2-6} & CAPs (ours) & 100 (0) & 100 & N/A & N/A \\
		
		\hline \hline \multirow{2}{*}{\specialcell{Collision,\\Heading}} & GC-DQL & 4.5 (7.5) & 0 & 0 & 6.9 (1.0) \\
		\cline{2-6} & CAPs (ours) & \textbf{98 (0)} & \textbf{80} & \textbf{80} & 6.9 (0.3) \\
		
		\hline \hline \multirow{2}{*}{\specialcell{Collision,\\Heading,\\Door}} & GC-DQL & 5.5 (3.1) & 0 & 0 & 7.0 (2.0) \\
		\cline{2-6} & CAPs (ours) & \textbf{98 (0)} & \textbf{80} & \textbf{80} & \textbf{8.1 (0.2)} \\
		\hline
	\end{tabular}
	}
	\caption{Comparison of our CAPs approach on a real-world RC car for the task of navigating a 75m loop and going near doors, which simulates a package delivery application. Our approach is able to successfully accomplish the task (bottom row), while the prior method does not succeed and can only accomplish a simple collision-avoidance task (top row). Note that only one model was trained with CAPs, and each experiment merely required changing the reward function at test time; in contrast, the prior method had to be retrained for each task. Additionally, our approach tasked with going towards doors (bottom row), compared to our approach not tasked with going towards doors (middle row), is indeed able to navigate while moving towards doors, showing our approach can learn from visual detection systems.}
	\label{table:exp-rccar}
	\vspace*{-10pt}
\end{wraptable}

We evaluated our approach on navigating a 75m loop in the environment, which contained five intersections. Two seconds before each intersection, a heading command was given to indicate which way to turn at the intersection. Fig.~\ref{fig:exp-rccar-tpv-fpv} shows images of our approach navigating the loop, while Table~\ref{table:exp-rccar} provides the corresponding quantitative results, with five trials per experiment. Our CAPs policy is able to successfully navigate this loop, while GC-DQL consistently crashes early on. However, DQL is able to successfully perform collision avoidance down a straight hallway (Table~\ref{table:exp-rccar}), indicating that its failure at the full task is due to its difficulty in learning from a multi-objective, goal-conditioned reward function.

In addition, we evaluated CAPs without the door subreward to determine if our approach could learn to predict and act on visual event cues. Note that we used the same CAPs model, but simply changed the reward function at test time. Table~\ref{table:exp-rccar} shows that the car sees 17\% more doors when the door subreward is included, which is a significant increase considering that the majority of the floor does not contain doors. This evaluation highlights the flexibility (\Qflexible) of our approach: using the same CAPs model, we can perform different tasks at test time.



\vspace*{-5pt}
\section{Discussion}
\vspace*{-5pt}

We presented CAPs, a general framework for flexible learning-based control. CAPs predicts event cues, which can be trained from off-policy data and flexibly combined at test time to accomplish various tasks. These event cues are automatically labeled using learned detection models, such as computer vision systems, which enable CAPs to be learned fully autonomously. We demonstrated CAPs on simulated and real-world robot navigation tasks, showing that it was indeed able to learn from these automatically labelled event cues and could flexibly accomplish various tasks at test time. Although CAPs can flexibly recombine its events into new tasks at test-time, CAPs can only learn about event cues within its predictive horizon. Investigating methods of combining CAPs with bootstrap-based methods~\cite{SuttonBarto} could lead to a new class of flexible, long-horizon learned controllers. The autonomous aspect of CAPs was enabled by leveraging existing detection models, for which we assumed the output of these detectors was the ground-truth; however, learned models are inherently fallible. Investigating how CAPs could be extended to cope with uncertain event cue labels is an important direction for real-world robot learning.



\acknowledgments{This work was supported by the National Science Foundation through IIS-1614653, the DARPA Assured Autonomy program, ARL DCIST CRA W911NF-17-2-0181, and Berkeley DeepDrive. Gregory Kahn was supported an NSF Graduate Research Fellowship.}


\bibliography{2018_CoRL_robonav}  


\newpage
\clearpage
\appendix

\setcounter{figure}{0}
\renewcommand{\thefigure}{S\arabic{figure}}

\section{Experiments}

For our CAPs model, a horizon of $H=16$ was used for the simulated forest, while the horizon was $H=12$ for the remaining experiments.

\textbf{Simulated forest.} The simulator is built on the Bullet physics engine~\citep{coumans2013bullet} and uses Panda3d~\citep{goslin2004panda3d} for image rendering. The robot's state $\bs \in \mathbb{R}^{6915}$ consists of sensor readings that could only be obtained from onboard sensors: $64 \times 36$ RGB images from a camera, forward speed readings from a wheel encoder, heading from a magnetometer, and collision readings from a bumper sensor. The robot's actions $\ba \in \mathbb{R}^2$ consist of the steering angle and setpoint for a speed PID controller. The robot selects and executes actions every $0.25$ seconds, although the simulator is still stepped at full resolution.

For the GC-DQL, the reward function used was
\begin{align*}
R(\bs_t, \ba_t) =
\begin{cases}
-500 & \text{if collision} \\
\cos(\bs_t^{(heading)} - \textsc{goal\_heading}) - 1 - \big( \frac{\bs_t^{(speed)} - \textsc{goal\_speed}}{\textsc{goal\_speed}} \big)^2 & \text{else}
\end{cases}.
\end{align*}
For GC-DQL-sep, the same subrewards were used to train each Q-function, and were then linearly combined to form the total Q-function using the same coefficients as the GC-DQL reward.

\textbf{Simulated city.} We used the robot car simulator CARLA~\cite{Dosovitskiy2017_CoRL}, a simulator with realistic renderings and physics. The robot's state $\bs \in \mathbb{R}^{5009}$ consists of sensor readings that could only be obtained from onboard sensors: $100 \times 50$ grayscale images from a camera, forward speed, heading, accelerations, and collision indicators. The robot's actions $\ba \in \mathbb{R}^2$ consisted of the steering angle and motor torque. The robot selects and executes actions every 0.25 seconds, although the simulator is still stepped at full resolution.

For GC-DQL, the reward function used was
\begin{align*}
R(\bs_t, \ba_t) =
\begin{cases}
-1000 & \text{if collision} \\
-\frac{ | \bs_t^{(speed)} - \textsc{goal\_speed} | }{\textsc{goal\_speed}} + \\ \;\;\;\; 10 \cdot \bs_t^{(lane\_seen)} \cdot (1 - | \bs_t^{(lane\_diff)} | ) - \\ \;\;\;\; \frac{5}{\pi} \cdot |\bs_t^{(heading)} - \textsc{goal\_heading} | & \text{else}
\end{cases}.
\end{align*}
For GC-DQL-sep, the same subrewards were used to train each Q-function, and were then linearly combined to form the total Q-function using the same coefficients as the GC-DQL reward.

\textbf{Real-world indoors.} The robot's state $\bs \in \mathbb{R}^{36867}$ consists of readings from the onboard sensors: $128 \times 96$ RGB images from a camera, forward speed from magnetic wheel encoders, collision indication from a bumper sensor, and heading from a magnetometer. The robot's action $\ba \in \mathbb{R}^1$ consists of the steering angle, while the speed is held fixed at $0.5$m/s. The robot selects and executes actions every 0.25 seconds.

For GC-DQL, the reward function used was
\begin{align*}
R(\bs_t, \ba_t) =
-\bs_t^{(coll)} + \frac{1}{\pi} \cdot (1 - | \bs_t^{(heading)} - \textsc{goal\_heading} |) + (1 - \bs_t^{(door\_frac)})
\end{align*}

\end{document}